\documentclass[11pt]{article}
\usepackage{amsmath}
\usepackage{amsfonts}

\usepackage[preprint]{acl}

\usepackage{times}
\usepackage{latexsym}
\usepackage{booktabs}
\usepackage{multirow}

\usepackage{authblk}

\usepackage[T1]{fontenc}

\usepackage[utf8]{inputenc}

\usepackage{microtype}

\usepackage{inconsolata}

\usepackage{graphicx}

\title{CriticGen: Generation-Aware Evaluation as Actionable Feedback}



\author[1]{Huifang Du\thanks{Equal contribution.}}   
\author[2]{Zecheng Zuo\protect\footnotemark[1]}     
\author[2]{Sen Wang}
\author[3]{Chenghao Fan}
\author[4]{Haofen Wang} 
\author[1]{Yehui Yang\thanks{Corresponding author (yangyehuisw@126.com). Also contactable via duhuifang123@gmail.com.}} 

\affil[1]{AI Lab, Qifu Technology, Beijing, China}
\affil[2]{Beijing University of Posts and Telecommunications, Beijing, China}
\affil[3]{University of Science and Technology Beijing, Beijing, China}
\affil[4]{Tongji University, Shanghai, China} 

\begin{document}
\maketitle
\begin{abstract}
Current evaluation methods for large language models are coarse-grained and decoupled from generation, producing generic explanations that fail to provide actionable feedback for model improvement.
We propose CriticGen, a fine-grained, generation-aware evaluation framework that turns evaluation into actionable control for answer improvement. CriticGen first generates sample-specific evaluation dimensions and scoring criteria under high-level categories such as subjective, objective, and self-derived constraints. These criteria then serve as a dynamic rubric for jointly producing a score, a reason, an executable refinement suggestion, and a refined answer. This rubric-conditioned refinement process enables models to diagnose flaws and perform targeted answer improvement.
Experimental results show that fine-grained evaluation should be both instance-specific and actionable. CriticGen induces higher-quality rubrics, improving relevance/coverage from 3.33/4.03 to 3.97/4.24. CriticGen also achieves the best score correlations, with 0.9556 Pearson and 0.9560 Spearman, and raises the F1 of criterion-grounded reasons and executable suggestions from 0.6369/0.5994 to 0.7554/0.7900. Crucially, its feedback translates into reliable answer improvement, improving 73.17\% of answers with a 93.28\% non-degradation rate.
\end{abstract}

\section{Introduction}\label{sec:intro}

\begin{figure*}[t]
  \centering
  \includegraphics[width=\textwidth]{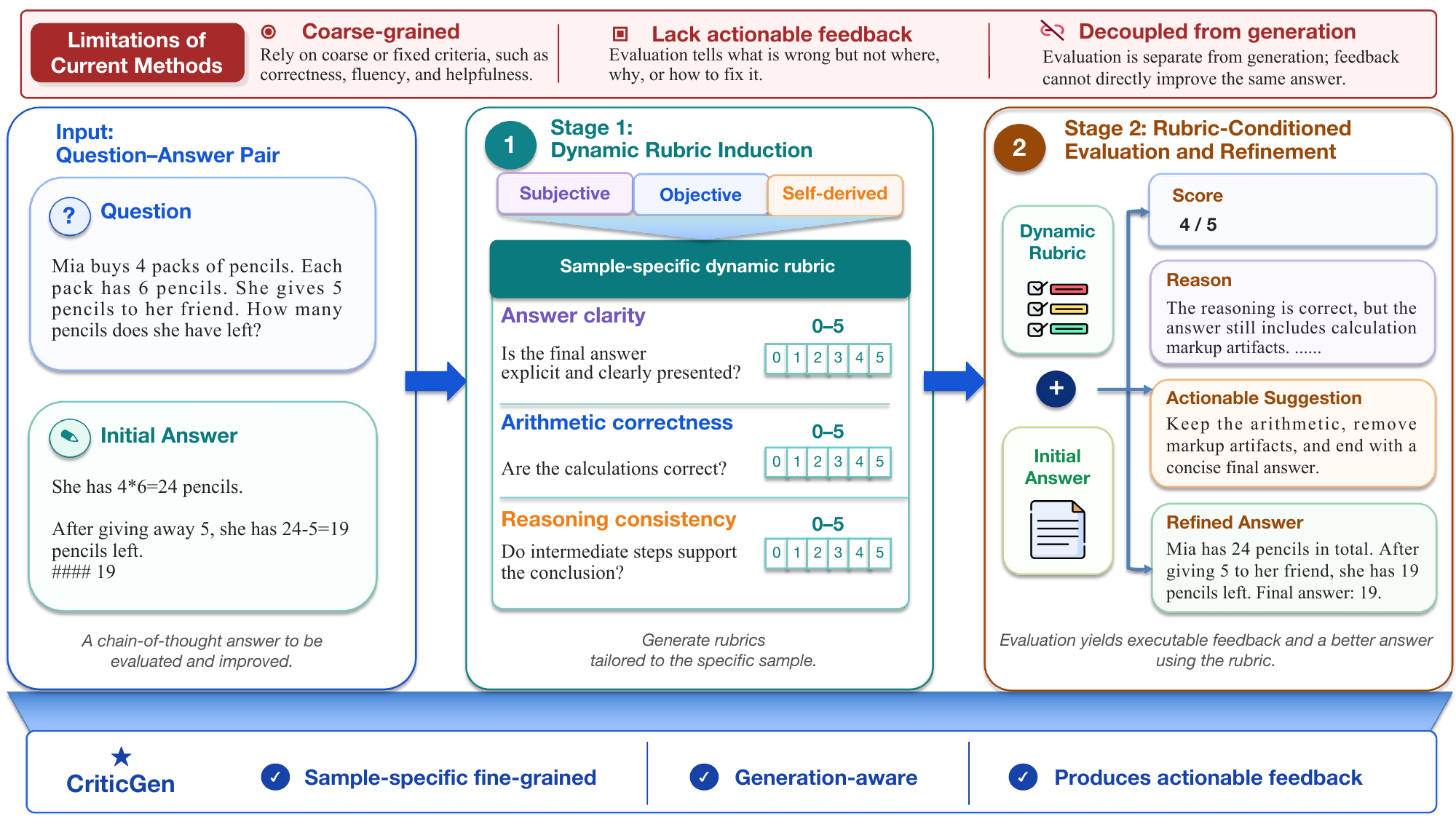}
  \caption{Illustration of CriticGen's core idea. Existing methods often provide fixed or generic feedback that is weakly connected to generation. CriticGen induces a sample-specific rubric and uses it to score, diagnose, plan, and refine the answer, turning evaluation into actionable control for answer improvement.}
  \label{fig:teaser}
\end{figure*}

Large language models (LLMs) are increasingly expected to produce not only correct final answers, but also faithful, consistent, and constraint-aware reasoning processes~\citep{wei2022chain,cobbe2021gsm8k}.
For chain-of-thought (CoT) outputs~\citep{wei2022chain}, evaluation is therefore no longer only a leaderboard metric.
It also serves as a diagnostic tool for identifying reasoning failures, curating training data, and guiding model improvement~\citep{ouyang2022training}.
However, current evaluation methods still leave a gap between \emph{judging} an answer and \emph{improving} it.

The first limitation is that many evaluation protocols are insufficiently adaptive to individual answers.
Outcome-level metrics such as exact match or shallow text overlap only indicate whether the final answer matches a reference, but do not reveal which reasoning aspect is weak~\citep{papineni2002bleu}.
Recent LLM-as-a-judge and fine-grained evaluation methods improve this by comparing responses, assigning ratings, or scoring outputs along multiple skills or criteria~\citep{zheng2023judging,liu2023geval,ye2024flask,kim2024prometheus}.
Generative judges further enhance interpretability by producing natural-language critiques together with judgments~\citep{li2024autoj}.
Nevertheless, their rubrics are often fixed at the task or scenario level.
Such criteria may cover general quality dimensions, but they may fail to capture the specific omissions, contradictions, or constraint violations that appear in a particular question--answer pair.

The second limitation is that evaluation signals are often not directly executable for generation.
Evaluator models such as Prometheus and LLaVA-Critic can provide fine-grained scores or feedback, and these signals can be used for reward-guided selection or later policy optimization~\citep{kim2024prometheus,xiong2025llavacritic,schulman2017proximal}.
This paradigm is effective, but it typically separates evaluation from generation: the evaluator is trained first, and its signal affects generation only through an additional optimization or selection stage.
Recent critic-based generation methods show that critique can improve model outputs, for example by refining erroneous reasoning steps or by training stronger policy behavior from critic data~\citep{zheng2025criticcot,wang2025llavacriticr1}.
Yet the critique is still not necessarily grounded in the same fine-grained criteria used for scoring, nor is it explicitly converted into a concrete edit plan for revising the same answer.

We ask whether evaluation can be made both \emph{sample-specific} and \emph{actionable}.
To this end, we propose \textbf{CriticGen}, a generation-aware evaluation framework that converts fine-grained evaluation into controllable answer refinement.
CriticGen contains two separately trained SFT components coupled by a rubric interface.
First, a rubric generator induces a sample-specific rubric \(R(q,a)=\{(d_i,C_i)\}_{i=1}^{m}\), where each dimension \(d_i\) is paired with a graded scoring criterion \(C_i\).
The rubric is organized under high-level constraint families, including subjective quality, objective correctness, and self-derived reasoning constraints, while remaining specific to the current question and answer.
Second, a rubric-conditioned evaluation-refinement model takes each \((q,a,d_i,C_i)\) as input and generates a structured trajectory \((s_i,r_i,u_i,a'_i)\): a score, a criterion-grounded reason, an executable refinement suggestion, and a refined answer.
In this trajectory, the reason explains \emph{why} the answer receives the score, while the suggestion specifies \emph{what} should be changed before rewriting.
Thus, CriticGen uses the same rubric to connect what to evaluate, why the answer is flawed, how it should be revised, and how the revised answer should be generated.

We construct two supervision datasets to support this framework.
\textsc{RubricData} trains the rubric generator to produce answer-specific dimensions and graded criteria.
\textsc{RefineData} trains the evaluation-refinement model to produce the complete score--reason--suggestion--rewrite trajectory under each rubric item.
Unlike reward-modeling or preference-optimization pipelines, CriticGen does not rely on a separate reinforcement-learning stage.
Instead, it studies whether structured supervised trajectories can directly teach models to transform evaluation into an executable refinement signal.

Experiments validate the central hypothesis of CriticGen: fine-grained evaluation should be both instance-specific and actionable.
Human evaluation shows that CriticGen induces higher-quality rubrics than a static checklist, improving relevance/coverage from 3.33/4.03 to 3.97/4.24.
On \textsc{RefineData}, CriticGen-Qwen3.5-9B achieves the best score correlations, reaching 0.9556 Pearson and 0.9560 Spearman, outperforming both the strongest open-weight baseline and GPT-5.
It also improves the generative components of evaluation, raising the best baseline F1 of criterion-grounded reasons and executable suggestions from 0.6369/0.5994 to 0.7554/0.7900.
More importantly, CriticGen turns feedback into reliable answer refinement, improving 73.17\% of answers while maintaining a 93.28\% non-degradation rate.
Out-of-domain results on Feedback Bench further show consistent transfer of rubric-conditioned evaluation ability.

We summarize our contributions as follows:
\begin{itemize}
 \item We propose sample-specific rubric-conditioned evaluation and refinement, which induces fine-grained rubrics for each question--answer pair and uses them to guide a score--reason--suggestion--rewrite trajectory, turning evaluation from descriptive judgment into actionable control for answer improvement.
  \item We construct RubricData and RefineData to train the two SFT components of CriticGen, enabling rubric induction and rubric-conditioned refinement without a separate reward-optimization stage.
  \item We demonstrate strong empirical performance across rubric quality, score correlation, reason and suggestion generation, answer refinement, and ablation studies, validating the effectiveness of rubric-conditioned refinement.
\end{itemize}

\section{Data Construction}\label{sec:data}

CriticGen requires supervision for two related but distinct capabilities: generating instance-specific rubrics and refining answers under a given rubric.
We therefore construct the training data in two layers, both starting from a diverse pool of candidate answers.

\paragraph{Diverse answer pool.}
We collect questions from five reasoning-oriented datasets: GSM8K~\citep{cobbe2021gsm8k}, LIMO~\citep{ye2025limo}, NaturalReasoning~\citep{yuan2025naturalreasoning}, NuminaMath-CoT~\citep{numina2024numinamath}, and QwQ-LongCoT-130K~\citep{amphora2024qwqlongcot}.
Together, these datasets cover arithmetic reasoning, symbolic/logical reasoning, open-domain natural-language reasoning, competition-level math, and long-form CoT generation, which helps the rubric and refinement models avoid overfitting to a single reasoning style.
For each question, we prompt generator models of different scales and capabilities, ranging from weaker open-weight models to stronger teacher models, to produce multiple candidate answers.
This multi-model generation is deliberate: using models with different error profiles leads to a more balanced distribution of data across the full 0--5 quality range.

\paragraph{Rubric supervision generation.}
For each question--answer pair \((q,a)\), GPT-5~\citep{openai2025gpt5} generates a sample-specific rubric
\(R^{*}=\{(d_i^{*},C_i^{*})\}_{i=1}^{m}\), where each item contains an evaluation
dimension and its 0--5 scoring criterion.
The dimensions are organized under the three high-level families used in our method:
subjective quality, objective correctness, and self-derived constraints.
We store each annotated instance as \((q,a,R^{*})\), which provides supervision for
rubric induction.

\paragraph{Evaluation-refinement supervision generation.}
Given each rubric item \((d_i^{*},C_i^{*})\), GPT-5 further acts as a judge and refiner.
It annotates a score \(s_i^{*}\), a criterion-grounded reason \(r_i^{*}\), an executable
refinement suggestion \(u_i^{*}\), and a refined answer \({a_i'}^{*}\).
Each resulting instance is stored as
\((q,a,d_i^{*},C_i^{*},s_i^{*},r_i^{*},u_i^{*},{a_i'}^{*})\), forming the supervision
for rubric-conditioned evaluation-to-refinement learning.
The score and reason capture how the answer satisfies the criterion, while the
suggestion and refined answer instantiate how the evaluation is converted into a
concrete improvement.

\paragraph{Score coverage completion.}
To reduce score imbalance, we check whether the answers for each question cover the full 0--5 quality range.
If some score regions are missing, we prompt GPT-5 to synthesize additional answers targeted at the missing quality levels.
For these synthesized answers, we repeat rubric supervision generation and evaluation-to-refinement supervision generation.
This step improves score coverage and increases answer diversity beyond what is obtained from the initial set of generator models.

\paragraph{Quality filtering.}
We first apply a consensus filter to acquire a reliable dataset in the evaluation-to-refinement stage.
Given the same question, answer, evaluation dimension, and scoring criteria, we ask Claude Sonnet 4.5~\citep{anthropic2025claudesonnet45},
Gemini 3.1 Pro Preview~\citep{google2026gemini31pro,google2026geminiapi} to independently generate the supervision.
We keep a sample only when their scores agree with the GPT-5 score, and discard samples with score disagreement.
We further apply a lexical-diversity and near-duplicate filter to avoid highly templated or repetitively phrased generations. Specifically, we compute field-level \(n\)-gram diversity and sample-level \(n\)-gram overlap over questions, answers, rubrics, reasons, refinement suggestions, and refined responses, and remove samples whose supervision is near-duplicated by previously retained instances. Appendix~\ref{app:ngram_filtering} provides the filtering details. The resulting data consist of two datasets: \textsc{RubricData} for dynamic
rubric induction, and \textsc{RefineData} for rubric-conditioned
evaluation-to-refinement learning. We split \textsc{RubricData} and
\textsc{RefineData} into training and test sets with ratios of 5:1 and
50:1, respectively. Table~\ref{tab:data_statistics} summarizes the final data
statistics.
\begin{table}[t]
  \centering
  \small
  \setlength{\tabcolsep}{3.2pt}
  \caption{
Statistics of the constructed datasets after quality filtering.
learning.
  }
  \label{tab:data_statistics}
  \begin{tabular*}{\columnwidth}{@{\extracolsep{\fill}}lcrrr@{}}
    \toprule
    Dataset & Train & Test & Total \\
    \midrule
    \textsc{RubricData} & 43{,}145 & 8{,}629 & 51{,}774 \\
    \textsc{RefineData} & 1{,}180{,}929 & 23{,}619 & 1{,}204{,}548 \\
    \bottomrule
  \end{tabular*}
\end{table}

\begin{figure*}[t]
  \centering
  \includegraphics[width=\textwidth]{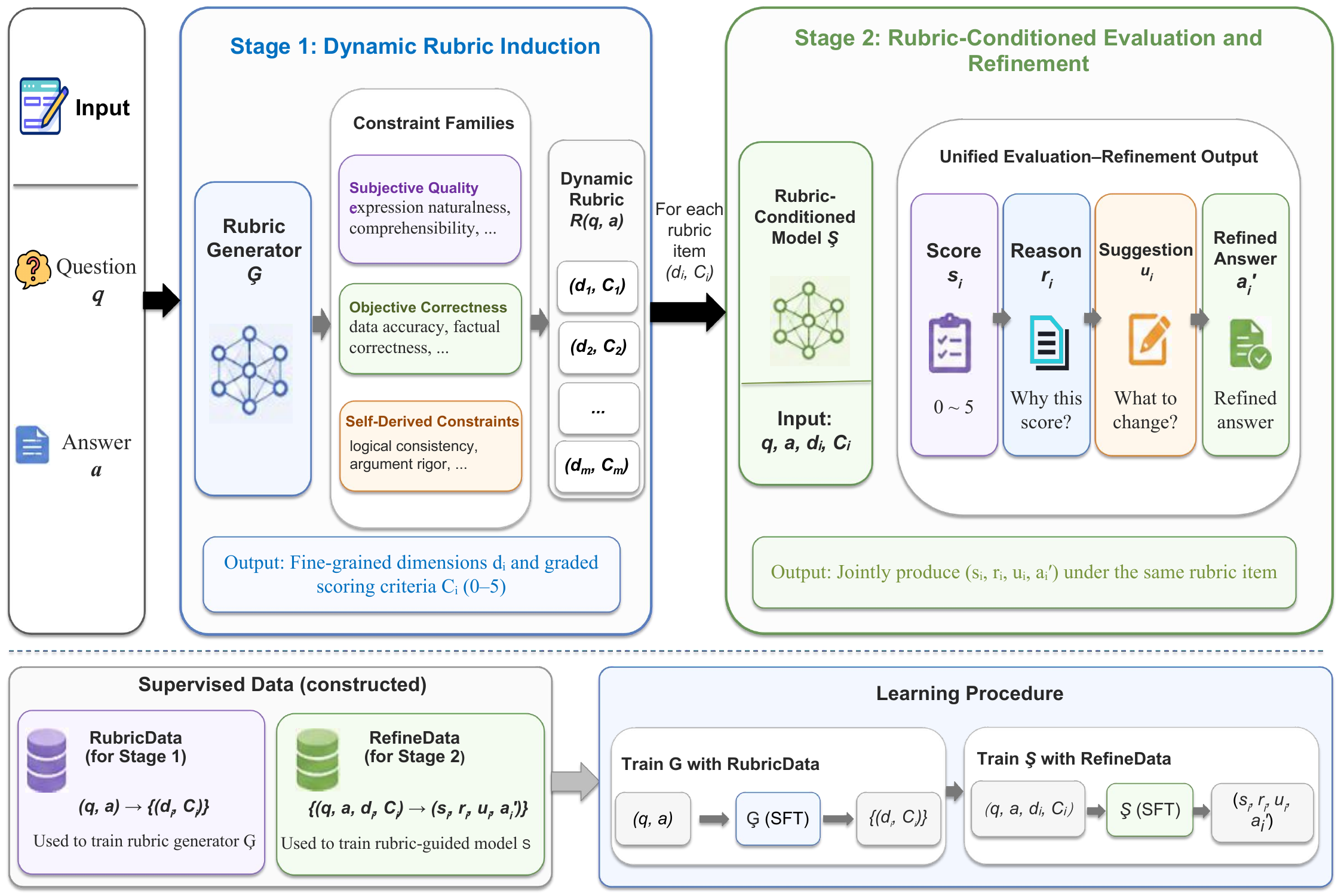}
  \caption{
Overview of the CriticGen framework. CriticGen first induces a sample-specific dynamic rubric for each question--answer pair, and then uses each rubric item to condition a unified model that jointly produces a score, a reason, an executable refinement suggestion, and a refined answer.
}
  \label{fig:overview}
\end{figure*}

\section{Method}\label{sec:method}

CriticGen is designed around the two limitations identified in the introduction: evaluation should be fine-grained for each instance, and its output should directly guide generation.
Accordingly, we separate the framework into two functions: \emph{rubric induction}, which specifies what should be evaluated, and \emph{rubric-conditioned refinement}, which uses the evaluation result to improve the same answer.

\subsection{Problem Formulation}
Let \(q\) denote a question and \(a\) a candidate reasoning answer.
Instead of learning only a scalar evaluator \(f(q,a)\rightarrow s\), CriticGen aims to
produce a structured set of rubric-conditioned evaluation and refinement outputs:
\begin{equation}
F(q,a) \rightarrow \{(d_i, C_i, s_i, r_i, u_i, a'_i)\}_{i=1}^{m}.
\end{equation}
Here, \(d_i\) is an evaluation dimension, \(C_i\) is its graded scoring criterion,
\(s_i\) is the predicted score, \(r_i\) is the score explanation, \(u_i\) is an
executable refinement suggestion, and \(a'_i\) is the refined answer.
This formulation makes evaluation not only descriptive, through \((s_i,r_i)\), but
also operational, through \((u_i,a'_i)\).

\subsection{Sample-Specific Rubric Induction}
The first component induces a dynamic rubric for each question--answer pair.
Given \((q,a)\), a rubric generator \(\mathcal{G}\) produces a set of dimensions and criteria:
\begin{equation}
R(q,a)=\{(d_i,C_i)\}_{i=1}^{m}.
\end{equation}
To ensure broad coverage while avoiding a fixed rubric, we organize dimensions under three high-level families:
\begin{itemize}
  \item \textbf{Subjective quality}: clarity, coherence, fluency, and readability;
  \item \textbf{Objective correctness}: factual accuracy, numerical correctness, and task-constraint satisfaction;
  \item \textbf{Self-derived constraints}: consistency among reasoning steps, absence of contradictions, and whether the conclusion follows from previous steps.
\end{itemize}
For each generated dimension \(d_i\), the model also produces a 0--5 criterion \(C_i\) that describes what different score levels mean for the current question and answer.
This differs from task-level or scenario-level rubrics: the generated criteria are tied to the concrete failure modes that may appear in the current answer.

\subsection{Rubric-Conditioned Evaluation and Refinement}
The second component takes a single rubric item \((d_i,C_i)\) and evaluates the candidate answer under that criterion.
For each \((q,a,d_i,C_i)\), a rubric-conditioned model \(\mathcal{S}\) generates four outputs:
\begin{equation}
\mathcal{S}(q,a,d_i,C_i) \rightarrow (s_i,r_i,u_i,a'_i).
\end{equation}
The score \(s_i\in\{0,\ldots,5\}\) measures how well the answer satisfies the criterion.
The reason \(r_i\) explains the score by identifying the relevant strength or weakness.
The refinement suggestion \(u_i\) turns this explanation into an executable instruction, such as adding a missing reasoning step, correcting a calculation, enforcing a neglected constraint, or clarifying a justification.
Finally, the refined answer \(a'_i\) is generated under the same criterion.

The key design choice is that \(u_i\) is not merely another explanation: \(r_i\) diagnoses the problem, while \(u_i\) specifies how the refiner should act.
We call \(u_i\) \emph{executable} because it gives a concrete edit operation and target content, e.g., ``insert a step before the final conclusion to apply the non-negativity constraint.''
Thus, the model learns to move from ``why this score'' to ``what should be changed'' and then to an improved answer.

\subsection{Training Objectives}
We train the two components with supervised fine-tuning on their corresponding
constructed supervision. This design makes the source of the gains explicit:
CriticGen is not optimized by a separate reward or preference stage, but learns
from two forms of structured supervision---instance-specific rubrics and
rubric-conditioned evaluation-to-refinement trajectories.

\paragraph{SFT for rubric induction.}
Let \(\mathcal{D}_{R}=\{(q^{(n)},a^{(n)},R^{*(n)})\}_{n=1}^{N_R}\)
denote \textsc{RubricData}, where
\(R^{*}=\{(d_i^{*},C_i^{*})\}_{i=1}^{m}\) is the constructed ground-truth rubric for a question--answer pair. The rubric generator
\(\mathcal{G}_{\theta}\) parameterizes a conditional distribution over
variable-length rubric items and is optimized with the empirical negative
log-likelihood:
\begin{equation}
\label{eq:sft_rubric}
L_{\mathrm{rubric}}(\theta)
=
-\mathbb{E}_{\mathcal{D}_{R}}
\log p_{\mathcal{G}_{\theta}}(R^{*}\mid q,a).
\end{equation}
In practice, this objective is implemented as a token-level cross-entropy over
the serialized dimensions and their 0--5 scoring criteria.

\paragraph{SFT for rubric-conditioned evaluation and refinement.}
Given \textsc{RefineData}, each training example contains an input
\(x_i=(q,a,d_i,C_i)\) and a consensus-filtered teacher output
\(y_i^*=(s_i^{*},r_i^{*},u_i^{*},a_i^{\prime *})\), where the score
\(s_i^{*}\) is retained only when multiple teacher models agree. We train the
rubric-conditioned model \(\mathcal{S}_{\phi}\) with a sequence likelihood objective over the serialized teacher output:
\begin{equation}
\label{eq:sft_refine}
L_{\mathrm{refine}}(\phi)
=
-\mathbb{E}_{\textsc{RefineData}}
\log p_{\mathcal{S}_{\phi}}(y_i^* \mid x_i).
\end{equation}
This single objective teaches the model to produce the complete
score--reason--revision-suggestion--rewrite trajectory under the same rubric
item. The fixed field order also provides lightweight structural supervision,
so the model learns not only the content of each field but also the interface
through which evaluation is converted into an executable refinement signal.

Although the two modules are trained independently, they are coupled at
inference time through the induced rubric, which serves as the interface between
evaluation target generation and actionable refinement.

\section{Experiments}\label{sec:experiments}
We structure our experiments around the two components of CriticGen: sample-specific rubric induction and rubric-conditioned evaluation and refinement.
For rubric induction, we examine whether the induced rubric
\(R(q,a)=\{(d_i,C_i)\}_{i=1}^{m}\)
serves as a useful assessment plan for a specific question--answer pair, using human judgments of rubric quality.
For rubric-conditioned evaluation and refinement, we test whether conditioning on a rubric item \((d_i,C_i)\) supports reliable scoring, criterion-grounded reasoning, actionable revision suggestions, and improved answer refinement.

\subsection{Setup}
\paragraph{Held-out data.}
For rubric induction quality, we sample 240 \((q,a)\) pairs, 48 from each of GSM8K, LIMO, NaturalReasoning, NuminaMath-CoT, and QwQ-LongCoT.
These pairs are from the rubric-induction test set of \textsc{RubricData}.
For \textbf{Ours}, we generate answer-specific rubrics \(R(q,a)=\{(d_i,C_i)\}\) (mean 8.4 dimensions per pair; Table~\ref{tab:rubric_quality}).
\textbf{Static} uses the same 240 \((q,a)\) pairs with nine fixed, instance-agnostic dimensions (three per family: subjective quality, objective correctness, self-derived constraints), keeping checklist length comparable to \textbf{Ours}.
This setup tests whether answer-specific rubrics improve relevance and coverage over static ones with matched dimension counts.

\paragraph{Human evaluation protocol.}
For each of the 240 held-out \((q,a)\) pairs, raters compare two rubrics under the same question and answer.
Rubric identities are blinded as Rubric~A/B.
Two independent raters with prior experience in LLM-as-a-judge evaluation and rubric design assign five-point Likert scores~\citep{likert1932technique} for two attributes:
\emph{relevance}, measuring how well the rubric dimensions and criteria fit the current \((q,a)\), and \emph{coverage}, measuring whether salient failure modes for judging the answer are represented.
If the two raters differ by at least one point on either rubric for either dimension, a senior adjudicator re-scores both rubrics; otherwise we average the two ratings.
Raters are not authors of this work.

\paragraph{Baseline models.}
We select baselines to cover backbone, scale, family, and proprietary-model
comparisons. Qwen2.5-7B-Instruct, Qwen3.5-4B, and Qwen3.5-9B are the backbones of our CriticGen variants, enabling direct task-specific fine-tuning. Qwen2.5-14B-Instruct
and Qwen2.5-32B-Instruct test whether gains come from scale alone, while
Mistral-7B-Instruct-v0.3 and Ministral-3-14B-Instruct-2512 provide non-Qwen
open-weight comparisons. GPT-5 serves as a high-capacity proprietary reference
evaluator.  Details and citations are provided in Appendix~\ref{app:baseline_models}.

\paragraph{Evaluation and Refinement Metric Protocol.}
For rubric-conditioned evaluation and refinement, we assess the outputs
\((s_i,r_i,u_i,a'_i)\) under each rubric item \((d_i,C_i)\) along four strands:

\textbf{Scores} \(s_i\): we report Pearson and Spearman correlations with the reference scores in \textsc{RefineData}.

\textbf{Reasons} \(r_i\) and \textbf{executable suggestions} \(u_i\): since both reasons and suggestions are free-form texts, we evaluate them by reference-based semantic-unit alignment rather than exact matching~\citep{zhang2020bertscore,sellam2020bleurt,li2025semantic_eval}. For a generated text \(x_i\) and its ground-truth reference \(x_i^*\), we split them into semantic units \(X_i=\{x_{i,j}\}_{j=1}^{m_i}\) and \(G_i=\{g_{i,k}\}_{k=1}^{n_i}\), and compute the pairwise similarity matrix
\begin{equation}
A^{(i)}_{k,j}=\mathrm{Sim}(g_{i,k},x_{i,j}).
\end{equation}
We then define recall-style coverage, precision-style consistency, and their harmonic mean as:
\begin{equation}
\begin{aligned}
R_i &=
\frac{1}{n_i}
\sum_{k=1}^{n_i}
\max_j A^{(i)}_{k,j}, \\
P_i &=
\frac{1}{m_i}
\sum_{j=1}^{m_i}
\max_k A^{(i)}_{k,j}, \\
\end{aligned}
\end{equation}
with \(F_i=2P_iR_i/(P_i+R_i)\). Here, \(R_i\) measures how well the generated text covers the ground-truth semantic units, while \(P_i\) measures whether the generated units are supported by the ground-truth reference. We report macro-averaged Precision, Recall, and F1 over all examples. This protocol is applied to reasons by setting \((x_i,x_i^*)=(r_i,r_i^*)\), and to executable suggestions by setting \((x_i,x_i^*)=(u_i,u_i^*)\).

\textbf{Refined answers} \(a'_i\):
a GPT-5 verifier scores both the original answer \(a\) and the refined answer \(a'_i\)
under the same rubric item \((d_i,C_i)\).
We report the improvement rate, unchanged rate, degradation rate, and the
non-degradation rate, where non-degradation means that the refined answer is
rated no worse than the original answer.

\subsection{Rubric Induction Quality}
\label{sec:exp_rubric_induction}

\paragraph{Rubric quality.}
Table~\ref{tab:rubric_quality} reports human judgments of rubric quality on 240 held-out \((q,a)\) pairs. The adjudication rate is 17.1\%, and the pre-adjudication Cohen's \(\kappa\)~\citep{cohen1968kappa} is 0.61 for relevance and 0.57 for coverage. Under the same \((q,a)\), raters assign higher scores to \emph{Ours} than to \emph{Static} on both relevance and coverage. The improvement is larger on relevance, increasing from 3.33 to 3.97, while coverage increases from 4.03 to 4.24.
This result suggests that induced rubrics mainly improve instance-level targeting rather than simply covering more generic evaluation aspects. Qualitatively, Static dimensions behave as templates aligned with broad failure families, so raters still see many salient checks covered and assign mid-to-high coverage.
Answer-conditioned induction adds criteria that explicitly mark omissions the fixed nine dimensions never single out, which is why raters score \emph{coverage} higher for \emph{Ours}.

\begin{table}[t]
  \centering
  \small
  \caption{Rubric induction: human judgments of rubric quality on 240 held-out \((q,a)\) pairs. Both rubrics are evaluated under the same question and answer; only the rubric text differs. Scores are post-adjudication means on a five-point Likert scale.}
  \label{tab:rubric_quality}
  \begin{tabular}{lccc}
    \toprule
    Rubric source & Rel. & Cov. & \#Dim. \\
    \midrule
    Static & 3.33 & 4.03 & 9.0 \\
    Ours & \textbf{3.97} & \textbf{4.24} & 8.4 \\
    \bottomrule
  \end{tabular}
\end{table}
Appendix~\ref{app:answer_sensitivity} further supports that the induced rubrics are answer-sensitive, not only higher-quality under human judgment.

\subsection{Rubric-Conditioned Evaluation Capability}

\begin{table}[t]
  \centering
  \small
  \setlength{\tabcolsep}{4.5pt}
  \caption{
Rubric-conditioned score evaluation quality.
P. and S. denote Pearson and Spearman correlations with reference scores in \textsc{RefineData}.
}
  \label{tab:evaluation_capability}
  \begin{tabular*}{\columnwidth}{@{\extracolsep{\fill}}lcc@{}}
    \toprule
    Model & P. & S. \\
    \midrule
    Qwen2.5-7B-Instruct & 0.6355 & 0.6300 \\
    Qwen2.5-14B-Instruct & 0.8144 & 0.8133 \\
    Qwen2.5-32B-Instruct & 0.8099 & 0.8210 \\
    Qwen3.5-4B & 0.7568 & 0.7504 \\
    Qwen3.5-9B & 0.8342 & 0.8328 \\
    Mistral-7B-Instruct-v0.3 & 0.6293 & 0.6236 \\
    Ministral-3-14B-Instruct-2512 & 0.8052 & 0.8108 \\
    GPT-5 & 0.9212 & 0.9143 \\
    \midrule
    CriticGen-Qwen2.5-7B & 0.9518 & 0.9521 \\
    CriticGen-Qwen3.5-4B & 0.9407 & 0.9413 \\
    CriticGen-Qwen3.5-9B & \textbf{0.9556} & \textbf{0.9560} \\
    \bottomrule
  \end{tabular*}
\end{table}

\begin{table}[t]
  \centering
  \small
  \setlength{\tabcolsep}{3.5pt}
  \caption{
Rubric-conditioned reason generation quality.
P., R., and F1 denote semantic-unit precision, recall, and F1 against reference reasons.
}
  \label{tab:reason_quality}
  \begin{tabular*}{\columnwidth}{@{\extracolsep{\fill}}lccc@{}}
    \toprule
    Model & P. & R. & F1 \\
    \midrule
    Qwen2.5-7B-Instruct & 0.6674 & 0.5190 & 0.5796 \\
    Qwen2.5-14B-Instruct & 0.6894 & 0.5098 & 0.5820 \\
    Qwen2.5-32B-Instruct & 0.6919 & 0.5106 & 0.5836 \\
    Qwen3.5-4B & 0.6636 & 0.5932 & 0.6218 \\
    Qwen3.5-9B & 0.6768 & 0.6092 & 0.6369 \\
    Mistral-7B-Instruct-v0.3 & 0.6638 & 0.5053 & 0.5701 \\
    Ministral-3-14B-Instruct-2512 & 0.6472 & 0.5865 & 0.6075 \\
    \midrule
    CriticGen-Qwen2.5-7B & 0.7150 & 0.7188 & 0.7150 \\
    CriticGen-Qwen3.5-4B & 0.7478 & 0.7560 & 0.7500 \\
    CriticGen-Qwen3.5-9B & \textbf{0.7549} & \textbf{0.7594} & \textbf{0.7554} \\
    \bottomrule
  \end{tabular*}
\end{table}

\begin{table}[t]
  \centering
  \small
  \setlength{\tabcolsep}{3.5pt}
 \caption{
Rubric-conditioned executable suggestion generation quality.
P., R., and F1 denote semantic-unit precision, recall, and F1 against reference suggestions.
}
  \label{tab:suggestion_quality}
  \begin{tabular*}{\columnwidth}{@{\extracolsep{\fill}}lccc@{}}
    \toprule
    Model & P. & R. & F1 \\
    \midrule
    Qwen2.5-7B-Instruct & 0.5988 & 0.4119 & 0.4844 \\
    Qwen2.5-14B-Instruct & 0.6540 & 0.4459 & 0.5260 \\
    Qwen2.5-32B-Instruct & 0.6921 & 0.4669 & 0.5533 \\
    Qwen3.5-4B & 0.6451 & 0.4740 & 0.5414 \\
    Qwen3.5-9B & 0.6976 & 0.5342 & 0.5994 \\
    Mistral-7B-Instruct-v0.3 & 0.6696 & 0.4553 & 0.5373 \\
    Ministral-3-14B-Instruct-2512 & 0.6372 & 0.5122 & 0.5612 \\
    \midrule
    CriticGen-Qwen2.5-7B & 0.7393 & 0.7357 & 0.7340 \\
    CriticGen-Qwen3.5-4B & 0.7783 & 0.7780 & 0.7754 \\
    CriticGen-Qwen3.5-9B & \textbf{0.7930} & \textbf{0.7921} & \textbf{0.7900} \\
    \bottomrule
  \end{tabular*}
\end{table}

Tables~\ref{tab:evaluation_capability}, \ref{tab:reason_quality}, and
\ref{tab:suggestion_quality} evaluate rubric-conditioned evaluation in terms of
score prediction, reason generation, and executable suggestion generation.
Overall, CriticGen consistently outperforms its backbone models and strong
instruction-following baselines, indicating that sample-specific rubrics provide
more effective evaluation signals than generic prompting.

For score prediction, Table~\ref{tab:evaluation_capability} shows that
CriticGen achieves much stronger agreement with reference scores.
CriticGen-Qwen3.5-9B obtains the best Pearson/Spearman correlations of
0.9556/0.9560, surpassing the strongest open-weight baseline Qwen3.5-9B
(0.8342/0.8328) by 0.1214/0.1232 and GPT-5 (0.9212/0.9143) by 0.0344/0.0417.
The gains are also evident for smaller models: CriticGen-Qwen2.5-7B improves
Qwen2.5-7B-Instruct from 0.6355/0.6300 to 0.9518/0.9521, suggesting that the
benefit comes from rubric-conditioned evaluation rather than model scale alone.

Tables~\ref{tab:reason_quality} and \ref{tab:suggestion_quality} further show
that CriticGen improves the generative side of evaluation. For reasons,
CriticGen-Qwen3.5-9B increases the best baseline F1 from 0.6369 to 0.7554, with
recall rising from 0.6092 to 0.7594. For executable suggestions, the gain is
larger, improving F1 from 0.5994 to 0.7900 and recall from 0.5342 to 0.7921.
Similar improvements appear on smaller backbones. These results support our
claim that evaluation should go beyond scalar scoring: by jointly modeling
scores, reasons, and executable suggestions under dynamic rubrics, CriticGen
turns evaluation into actionable control for answer refinement.

\subsection{Refinement Capability}
\label{sec:exp_refinement_capability}

\begin{figure}[t]
  \includegraphics[width=\columnwidth]{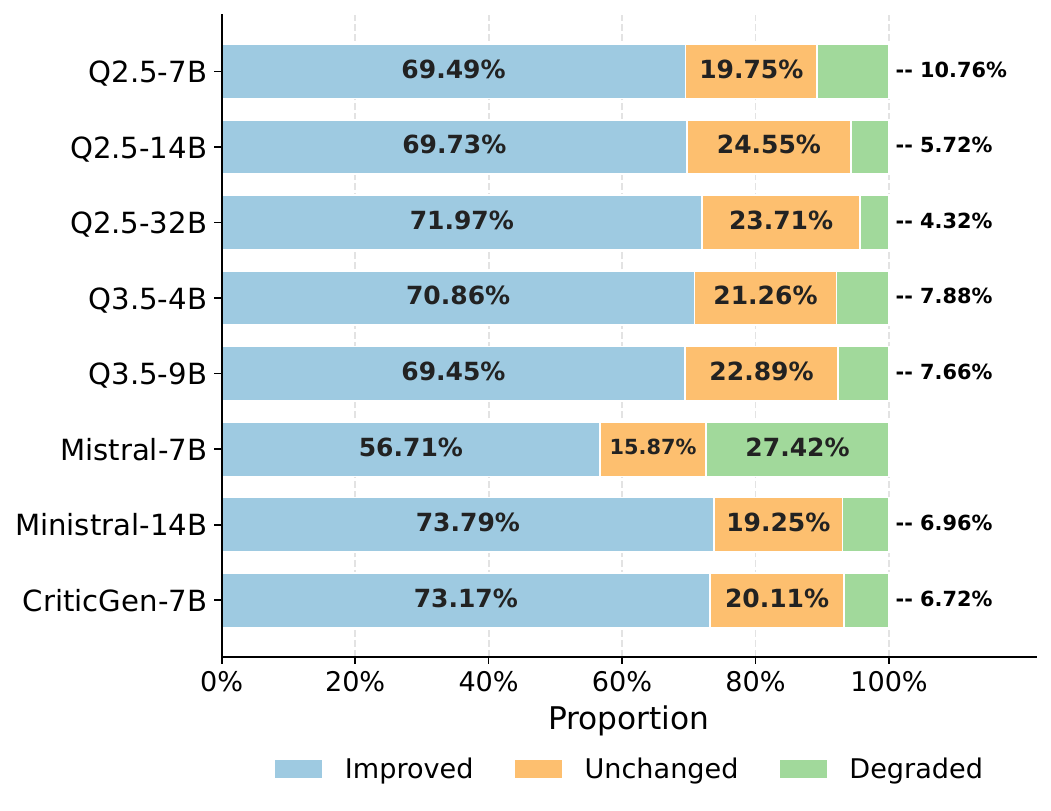}
  \caption{
Rubric-conditioned refinement outcomes.
Each bar shows the proportions of improved, unchanged, and degraded cases
after refinement under the same rubric item.
}
  \label{fig:refined_answer_outcomes}
\end{figure}

Figure~\ref{fig:refined_answer_outcomes} compares each rubric-conditioned
refined answer \(a'_i\) with the original answer \(a\) under the same rubric
item, and categorizes the outcome as improved, unchanged, or degraded. The
non-degradation rate is the sum of improved and unchanged cases.

Overall, most models improve many answers, but their reliability varies.
Mistral-7B-Instruct-v0.3 performs worst, with 56.71\% improved cases and
27.42\% degraded cases. Stronger baselines are more stable:
Qwen2.5-32B-Instruct achieves the highest non-degradation rate of 95.68\%,
while Ministral-3-14B-Instruct-2512 obtains the highest baseline improvement
rate of 73.79\%. CriticGen achieves a strong balance, improving 73.17\% of
cases, leaving 20.11\% unchanged, and degrading only 6.72\%, for a
non-degradation rate of 93.28\%. Compared with its Qwen2.5-7B backbone,
CriticGen improves both the improvement rate, from 69.49\% to 73.17\%, and the
non-degradation rate, from 89.24\% to 93.28\%.

These results show that CriticGen converts fine-grained evaluation into
actionable refinement signals, producing revisions that improve answers while
keeping degradation risk low.

\subsection{Ablation Study on Evaluation-to-Refinement}
\label{sec:ablation_refinement}

We ablate the structured score--reason--suggestion--rewrite trajectory to study
its role in rubric-conditioned refinement. Since rubric induction is evaluated
separately in Section~\ref{sec:exp_rubric_induction}, all variants use the same
rubric condition \((d_i,C_i)\) and differ only in the second-stage supervision
fields. All experiments are based on CriticGen-Qwen3.5-4B-SFT and evaluated on
the held-out split of \textsc{RefineData}.

Table~\ref{tab:ablation_refinement} shows that Full CriticGen performs best,
achieving an improved rate of 73.54. Removing reason or suggestion supervision
only mildly affects the remaining evaluation metrics, but reduces refinement
success to 60.48 and 61.57, respectively. The rewrite-only variant also
underperforms Full CriticGen (62.35) and loses interpretable score, reason, and
suggestion outputs. These results indicate that criterion-grounded diagnosis
and executable edit planning are crucial for converting evaluation into
effective and controllable refinement.

\begin{table}[t]
  \centering
  \small
  \setlength{\tabcolsep}{2.8pt}
  \caption{
Ablation study on the second-stage evaluation-to-refinement model.
All variants use the same rubric condition \((d_i,C_i)\).
Score P. denotes Pearson correlation for score prediction; reason and suggestion
quality are measured by semantic-unit F1. Improved denotes the proportion of
refined answers rated better than the original answers.
  }
  \label{tab:ablation_refinement}
  \begin{tabular*}{\columnwidth}{@{\extracolsep{\fill}}lcccc@{}}
    \toprule
    Variant & Sc. P. & R. F1 & Sugg. F1 & Improved \\
    \midrule
    Full CriticGen
    & \textbf{0.9407}
    & \textbf{0.7500}
    & \textbf{0.7754}
    & \textbf{73.54} \\
    w/o reason
    & 0.9323
    & --
    & 0.7663
    & 60.48 \\
    w/o suggestion
    & 0.9304
    & 0.7484
    & --
    & 61.57 \\
    rewrite only
    & --
    & --
    & --
    & 62.35 \\
    \bottomrule
  \end{tabular*}
\end{table}

\section{Conclusion}\label{sec:conclusion}
We presented CriticGen, a generation-aware evaluation framework that turns fine-grained evaluation into actionable control for answer refinement. CriticGen first induces sample-specific rubrics for each question-answer pair and then uses the induced criteria to jointly produce a score, a criterion-grounded reason, an executable refinement suggestion, and a refined answer. Experiments show that CriticGen improves rubric quality, score correlation, reason and suggestion generation, and answer refinement reliability. These results suggest that evaluation can serve not only as a diagnostic signal but also as a direct mechanism for controllable generation improvement.

\section*{Limitations}
\label{sec:limitations}

This work has several limitations. First, CriticGen is trained with supervised fine-tuning, which depends on the quality and coverage of the constructed supervision data. Although our results show strong performance, the framework may still inherit biases or blind spots from the generated rubrics and refinement trajectories. Second, our current experiments focus on text reasoning tasks, and further validation is needed on broader domains such as multimodal reasoning, long-form generation, and interactive decision-making. Third, while CriticGen directly converts evaluation into refinement through a structured SFT trajectory, it does not yet optimize the model with reinforcement learning. Future work can further use reinforcement learning or preference optimization to strengthen the alignment between rubric-conditioned feedback and actual answer improvement.

\bibliography{custom}

\appendix
\section{Related Work}\label{sec:related}

\paragraph{Reasoning Evaluation Beyond Final Answers.}
Chain-of-thought prompting improves mathematical and multi-step reasoning by eliciting intermediate steps~\citep{wei2022chain,cobbe2021gsm8k}, but also complicates evaluation: correct answers may still contain incomplete, unclear, or constraint-inconsistent reasoning. Our work evaluates and revises reasoning under explicit criteria, rather than merely encouraging longer chains.

\paragraph{LLM Judges and Fine-Grained Criteria.}
LLM-as-a-judge methods scale automatic evaluation by prompting or training models to compare responses and assign ratings~\citep{zheng2023judging,liu2023geval,wang2024pandalm}. Fine-grained evaluators decompose quality into skills or criteria~\citep{ye2024flask,kim2024prometheus}, and generative judges further produce critiques or feedback~\citep{li2024autoj,wang2023shepherd}. However, their criteria are often task- or scenario-level, and their outputs usually remain scores, critiques, or rewards. CriticGen instead induces a sample-specific rubric for each question--answer pair and conditions both evaluation and refinement on it.

\paragraph{Evaluation as Reward for Model Improvement.}
Evaluators are also used as reward models or selection functions for improving separate policy models. Preference modeling and RLHF-style methods learn reward signals from human or model feedback~\citep{ziegler2019fine,ouyang2022training,schulman2017proximal}, while Prometheus and LLaVA-Critic provide fine-grained feedback for reward-guided selection or reinforcement learning~\citep{kim2024prometheus,xiong2025llavacritic}. In these methods, evaluation and generation are typically separated. CriticGen studies a more direct path: the model scores the current answer, diagnoses weaknesses, generates an executable refinement suggestion, and rewrites the answer under the same rubric.

\paragraph{Self-Critique and Critic-Based Generation.}
Self-refinement and Reflexion show that model-generated feedback can improve later outputs through critique and revision~\citep{madaan2023selfrefine,shinn2023reflexion}. Critic-CoT trains step-wise critiques for reasoning errors~\citep{zheng2025criticcot}, and LLaVA-Critic-R1 shows that critic training can yield strong policy behavior after reinforcement learning on critic data~\citep{wang2025llavacriticr1}. These works validate critic signals, while CriticGen focuses on converting dynamically induced, fine-grained rubrics into directly executable refinement signals by grounding the score, reason, suggestion, and rewrite in the same instance-specific criteria.

\section{Data Filtering}
\label{app:ngram_filtering}

To ensure that the constructed data are not dominated by templated or repetitive generations, we conduct lexical-diversity analysis and near-duplicate filtering during data construction.
This process is applied after score-consensus filtering and before the final train/test split.

\paragraph{Text normalization.}
We normalize all textual fields before computing lexical statistics.
Specifically, we lowercase English text, remove redundant whitespace, and tokenize each field into word-level tokens.
For each instance, we consider both the input-side fields and the generated fields in the constructed data.
The input-side fields include the question and the original answer.
The generated fields include the induced rubric, the evaluation reason, the executable refinement suggestion, and the refined answer.
We analyze these fields separately because they play different roles in the constructed evaluation-to-refinement trajectory.

\paragraph{Field-level lexical diversity.}
We first measure field-level lexical diversity using Distinct-\(n\), defined as the ratio of unique \(n\)-grams to the total number of \(n\)-grams in a field collection:
\begin{equation}
\mathrm{Distinct}\text{-}n
=
\frac{
|\mathrm{UniqueNgrams}_n(\mathcal{T})|
}{
|\mathrm{AllNgrams}_n(\mathcal{T})|
},
\end{equation}
where \(\mathcal{T}\) denotes all texts from a specific field.
Higher Distinct-\(n\) values indicate greater lexical diversity and lower template reuse.
We compute Distinct-2, Distinct-3, and Distinct-4 for questions, answers, rubrics, reasons, executable suggestions, and refined answers.
Table~\ref{tab:ngram_diversity} reports the field-level diversity statistics.

\begin{table}[t]
\centering
\small
\begin{tabular}{lccc}
\toprule
Field & Distinct-2 & Distinct-3 & Distinct-4 \\
\midrule
Questions       & 0.132 & 0.232 & 0.410 \\
Answers         & 0.182 & 0.431 & 0.547 \\
Rubrics         & 0.258 & 0.362 & 0.569 \\
Reasons         & 0.243 & 0.432 & 0.651 \\
Suggestions     & 0.213 & 0.396 & 0.678 \\
Refined answers & 0.199 & 0.480 & 0.693 \\
\bottomrule
\end{tabular}
\caption{
Field-level lexical diversity of the constructed data.
The generated evaluation-to-refinement fields, especially reasons, executable suggestions, and refined answers, show substantial high-order \(n\)-gram diversity.
Questions exhibit lower diversity because they are more task-driven and often follow similar query forms.
Overall, the results suggest that the constructed data are not dominated by repeated lexical templates.
}
\label{tab:ngram_diversity}
\end{table}

\paragraph{Sample-level near-duplicate filtering.}
Field-level diversity measures the overall lexical variation of the dataset, but does not directly identify near-duplicate instances.
We therefore apply a sample-level overlap filter.
For each candidate instance, we concatenate its generated fields, including the rubric, reason, executable suggestion, and refined answer, into a single text sequence.
We then compute the 4-gram Jaccard overlap between the candidate and previously retained instances:
\begin{equation}
J_4(x_i, x_j)
=
\frac{
|\mathcal{G}_4(x_i) \cap \mathcal{G}_4(x_j)|
}{
|\mathcal{G}_4(x_i) \cup \mathcal{G}_4(x_j)|
},
\end{equation}
where \(\mathcal{G}_4(x)\) denotes the set of 4-grams in instance \(x\).
A candidate instance is removed if its maximum 4-gram Jaccard overlap with any retained instance exceeds \(\tau=0.75\).
In implementation, we use MinHash-based locality-sensitive hashing to retrieve high-overlap candidates efficiently, followed by exact Jaccard computation for final filtering.

Overall, Table~\ref{tab:ngram_diversity} shows that the constructed data retain meaningful lexical variation across both input-side and generated fields.
Although the questions are relatively more standardized due to the task-oriented data construction process, the generated reasons, executable suggestions, and refined answers exhibit high-order diversity, indicating that the evaluation-to-refinement trajectories are not merely repeated templates.

\section{Answer Sensitivity of Induced Rubrics}
\label{app:answer_sensitivity}

\paragraph{Data.}
We use 180 held-out questions with 3--5 candidate answers each. The questions are held out from training and are used only for the answer-sensitivity analysis.

\paragraph{Protocol.}
To measure whether rubrics are answer-sensitive, we compare rubrics generated for different answers to the same question.
For \textbf{Ours}, we compare answer-specific rubrics \(R(q,a)\) and \(R(q,a')\).
For \textbf{Q-only}, we use a shared question-only rubric \(R(q)\) for both answers, which cannot vary with the answer by construction.
For \textbf{Q-only$\times$2}, we compare two independently generated question-only rubrics for the same question, providing a lexical-noise control for repeated rubric generation.
Rubric change is measured by the lexical Jaccard shift over pooled \texttt{dimension\_name} tokens.

\paragraph{Results.}
We evaluate whether induced rubrics change when the answer changes.
As a lightweight, reproducible proxy, we measure \emph{lexical} turnover in \texttt{dimension\_name} strings only and exclude the 0--5 criterion text, which reduces sensitivity to long paraphrases in scoring anchors.
For a rubric \(R(q,a)\), we take every \texttt{dimension\_name}, lowercase it, tokenize it into alphanumeric word \emph{types}, and pool those types into a single set \(W_a\) (duplicate tokens across dimensions count once).
Fixing the same question \(q\), we induce \(R(q,a)\) and \(R(q,a')\) from distinct answers \(a\) and \(a'\), construct \(W_a\) and \(W_{a'}\), and compute the Jaccard distance
\begin{equation}
D_{\mathrm{Jaccard}}
=
1 -
\frac{|W_a \cap W_{a'}|}
{|W_a \cup W_{a'}|}.
\end{equation}
The distance is 0 if both token sets are empty.
Larger \(D_{\mathrm{Jaccard}}\) means less overlap in the word types appearing anywhere in dimension titles when the conditioning answer changes.
Table~\ref{tab:rubric_sensitivity} summarizes means on 180 held-out questions.
\textbf{Q-only} uses the same question-only rubric for both answers, so its shift is zero by design.
\textbf{Q-only$\times$2} compares two independently generated question-only rubrics and shows a small shift (0.09), reflecting lexical variation from repeated generation.
\textbf{Ours} compares \(R(q,a)\) and \(R(q,a')\) and achieves a much larger shift of 0.37.
This confirms a necessary property of sample-specific evaluation: the rubric changes with the answer, not just the question.
Together with the human evaluation in Section~\ref{sec:exp_rubric_induction}, these results indicate that our rubrics are both more relevant to each \((q,a)\) and more sensitive to answer-specific evaluation needs.

\begin{table}[t]
  \centering
  \small
  \caption{
Answer sensitivity of induced rubrics on 180 held-out questions. 
\emph{Q-only} uses a shared question-level rubric, \emph{Q-only$\times$2} compares two independently generated question-level rubrics, and \emph{Ours} compares answer-conditioned rubrics for different answers to the same question.
}
  \label{tab:rubric_sensitivity}
  \begin{tabular}{lcc}
    \toprule
    Comparison & $n$ & Jaccard shift \\
    \midrule
    Q-only & 180 & 0.00 \\
    Q-only$\times$2 & 180 & 0.09 \\
    Ours & 180 & \textbf{0.37} \\
    \bottomrule
  \end{tabular}
\end{table}

\section{Baseline Models and CriticGen Variants}
\label{app:baseline_models}

We compare CriticGen with a diverse set of instruction-following LLM baselines. 
Unless otherwise specified, all baseline models are evaluated under the same 
rubric-conditioned prompts and output format, and are not fine-tuned on our 
constructed data.

\begin{itemize}
    \item \textbf{Qwen2.5-7B-Instruct.}
    Qwen2.5-7B-Instruct serves as the primary open-weight backbone in our
    experiments~\citep{qwen2025qwen25}. It provides a strong 7B-scale
    instruction-following baseline and is used to assess the zero-shot or
    prompt-only capability of a model with the same architecture scale as our
    main CriticGen variant.

    \item \textbf{Qwen2.5-14B-Instruct.}
    We include Qwen2.5-14B-Instruct as a medium-scale Qwen2.5 baseline
    \citep{qwen2025qwen25}. Comparing it with Qwen2.5-7B-Instruct and
    Qwen2.5-32B-Instruct allows us to examine whether a larger model capacity alone improves rubric-conditioned evaluation and refinement.

    \item \textbf{Qwen2.5-32B-Instruct.}
    Qwen2.5-32B-Instruct is used as a larger Qwen2.5-family baseline
    \citep{qwen2025qwen25}. It provides a scale-controlled comparison for
    evaluating whether CriticGen's gains can be attributed to task-specific
    supervision rather than simply to increasing the number of model parameters.

    \item \textbf{Qwen3.5-4B.}
    We include Qwen3.5-4B as a newer small-scale Qwen-series baseline
    \citep{qwen2026qwen35_4b}. It allows us to test whether the recent base-model
    improvements can already support rubric-conditioned evaluation without
    task-specific fine-tuning.

    \item \textbf{Qwen3.5-9B.}
    Qwen3.5-9B is used as a newer medium-scale Qwen-series baseline
    \citep{qwen2026qwen35_9b}. Together with Qwen3.5-4B, it helps evaluate the
    robustness of CriticGen's training strategy across different Qwen
    generations and parameter scales.

    \item \textbf{Mistral-7B-Instruct-v0.3.}
    We use Mistral-7B-Instruct-v0.3 as a non-Qwen open-weight
    instruction-following baseline~\citep{jiang2023mistral7b,mistral2024mistral7binstructv03}.
    Its inclusion helps assess whether CriticGen's advantages are specific to the Qwen model family or remain competitive against independently developed
    open-weight instruction models.

    \item \textbf{Ministral-3-14B-Instruct-2512.}
    We include Ministral-3-14B-Instruct-2512 as a recent non-Qwen
    instruction-following baseline~\citep{mistral2025ministral314b}. It
    provides an additional comparison point for evaluating CriticGen against
    contemporary open-weight instruction models.

    \item \textbf{GPT-5.}
    GPT-5 is included as a strong proprietary model baseline
    \citep{openai2025gpt5,openai2025gpt5systemcard}. In addition to serving as
    a high-capability comparison model, it is also used as a reference judge in
    parts of our data construction and verification pipeline.
\end{itemize}

Our CriticGen variants are obtained by applying the same supervised fine-tuning
procedure to select Qwen backbones using our constructed training data.
Specifically, CriticGen-Qwen2.5-7B-SFT, CriticGen-Qwen3.5-4B-SFT, and
CriticGen-Qwen3.5-9B-SFT are initialized from Qwen2.5-7B-Instruct
\citep{qwen2025qwen25}, Qwen3.5-4B~\citep{qwen2026qwen35_4b}, and
Qwen3.5-9B~\citep{qwen2026qwen35_9b}, respectively, and trained on
\textsc{RubricData} and \textsc{RefineData}. This design isolates the effect of
CriticGen's data and learning objective from the choice of backbone: the base
instruction models measure zero-shot or prompt-only rubric-conditioned
evaluation ability, while the CriticGen variants measure the effect of learning
to generate sample-specific rubrics and to produce the coupled
score--reason--revision-suggestion--rewrite trajectory from our supervision.

\section{Training Details}
\label{app:training_details}

We train CriticGen variants with supervised fine-tuning (SFT).
All training runs are initialized from the corresponding instruction-tuned backbone and use the same serialized output format as described in Section~\ref{sec:method}.
For rubric induction, the model is trained to generate the complete set of induced rubric items, including dimension names and 0--5 scoring criteria.
For rubric-conditioned evaluation and refinement, the model is trained to generate the complete target sequence, including the score, reason, executable revision suggestion, and refined answer.
We optimize the standard next-token prediction loss with AdamW, using a learning rate of \(2\times10^{-5}\), a cosine learning-rate schedule, a warmup ratio of 0.03, and weight decay of 0.01.
Unless otherwise specified, we train for 3 epochs with bf16 precision, gradient checkpointing, and a maximum sequence length of 4096 tokens.
The effective batch size is set to 128 through gradient accumulation.

We choose the SFT checkpoint as the main CriticGen model in all reported experiments.
This choice keeps the empirical comparison focused on the proposed data construction and evaluation-to-refinement formulation.
It also avoids introducing an additional preference-optimization stage whose gains may depend on preference-pair construction, verifier calibration, and reward-model noise rather than on the core CriticGen framework.

\section{Out-of-Domain Evaluation on Feedback Bench}
\label{app:feedback_bench}

To further examine the out-of-domain generalization ability of CriticGen, we evaluate it on Feedback Bench from Prometheus~\citep{kim2024prometheus}. 
Feedback Bench is designed for absolute grading with customized score rubrics. 
Each instance provides an instruction, a response to be evaluated, a reference answer, and a fine-grained scoring rubric, and the evaluator LM is required to assign a scalar score and feedback according to the given rubric.

Different from our in-domain \textsc{RefineData}, Feedback Bench is not used during training and follows an independently constructed evaluation format. 
We therefore use it as an out-of-domain benchmark to test whether the evaluation capability learned by CriticGen transfers to external rubric-conditioned scoring scenarios.
To ensure a fair comparison, we compare CriticGen-Qwen2.5-7B-SFT with evaluator LMs of a similar scale, including Qwen2.5-7B-Instruct, Qwen3.5-4B, and Mistral-7B-Instruct-v0.3.

Following the evaluation protocol of Prometheus, we measure the correlation between model-predicted scores and reference scores using Pearson correlation, Spearman correlation, and Kendall's \(\tau\).
The results are shown in Table~\ref{tab:feedback_bench_ood}.

\begin{table}[t]
  \centering
  \small
  \setlength{\tabcolsep}{4.5pt}
  \caption{
Out-of-domain evaluation on Feedback Bench.
We report Pearson, Spearman, and Kendall's \(\tau\) correlations between model-predicted scores and reference scores.
  }
  \label{tab:feedback_bench_ood}
  \begin{tabular*}{\columnwidth}{@{\extracolsep{\fill}}lccc@{}}
    \toprule
    Model & Pearson & Spearman & Kendall \(\tau\) \\
    \midrule
    Qwen2.5-7B-Instruct & 0.7520 & 0.7522 & 0.6606 \\
    Qwen3.5-4B & 0.7373 & 0.7451 & 0.6532 \\
    Mistral-7B-Instruct-v0.3 & 0.6783 & 0.6801 & 0.5943 \\
    \midrule
    CriticGen-Qwen2.5-7B-SFT & \textbf{0.7556} & \textbf{0.7612} & \textbf{0.6755} \\
    \bottomrule
  \end{tabular*}
\end{table}

CriticGen-Qwen2.5-7B-SFT achieves the best performance across all three correlation metrics.
Compared with its base model Qwen2.5-7B-Instruct, CriticGen improves Pearson correlation from 0.7520 to 0.7556, Spearman correlation from 0.7522 to 0.7612, and Kendall's \(\tau\) from 0.6606 to 0.6755.
Although the improvement is modest, the consistent gains across all metrics suggest that CriticGen's rubric-conditioned evaluation capability can transfer to out-of-domain customized scoring settings.

\end{document}